\documentclass{article}

\usepackage{arxiv}

\usepackage[utf8]{inputenc} 
\usepackage[T1]{fontenc}    
\usepackage{hyperref}       
\usepackage{url}            
\usepackage{booktabs}       
\usepackage{amsfonts}       
\usepackage{nicefrac}       
\usepackage{microtype}      
\usepackage{lipsum}		
\usepackage{graphicx}
\usepackage{natbib}
\usepackage{doi}

\usepackage{booktabs}
\usepackage{multirow}
\usepackage{longtable}
\usepackage{amsmath}
\usepackage{amssymb}
\usepackage{enumitem}
\usepackage{algorithm}
\usepackage{algorithmic}
\newcommand{\cmark}{\checkmark}

\title{SpectraDINO: Modality-Conditioned Adaptation of RGB Vision Foundation Models Across Infrared Bands}

\author{Yagiz Nalcakan\thanks{Corresponding Author)}, Hyeongjin Ju, Incheol Park, Sanghyeop Yeo, Youngwan Jin, Shiho Kim\\
	College of Computing \\
	Yonsei University \\
	Incheon, Republic of Korea \\
	\texttt{$[ynalcakan,wngudwls000,incheol97,dutkdguq,thatnn,shiho]$@yonsei.ac.kr} \\
}

\date{}

\renewcommand{\shorttitle}{SpectraDINO}
  
\begin{document}
\maketitle

\begin{abstract}
	Vision foundation models (VFMs) pretrained on large-scale RGB data provide strong general-purpose representations, yet infrared perception, which is essential for robotics and driving in low light and adverse weather, still relies on backbones built per sensor. Existing transfer methods bridge the visible-infrared gap for only a single band or a single RGB-IR pair, leaving no unified backbone that spans the infrared spectrum. We present SpectraDINO, a modality-conditioned adaptation of an RGB vision foundation model that supports NIR, SWIR, and LWIR with a single shared ViT. All modality-specific behavior is conditioned on the active sensor via lightweight input stems, learned modality embeddings, and per-modality bottleneck adapters, while the transformer weights remain shared. We propose a staged protocol to distill the frozen DINOv2 teacher into a spectral student via cosine distillation, symmetric contrastive alignment, patch-level correspondence, and a queue-based neighborhood-preservation loss, and then selectively fine-tune the deeper blocks to adapt without forgetting RGB priors. Across seven detection and segmentation benchmarks spanning all three infrared bands, this single conditioned backbone matches or surpasses modality-specific methods, improving mIoU over the best published results by 2.6 points on FMB and 1.4 on SemanticRT, and exceeding identically fine-tuned foundation-model baselines by 2.7 mAP on SWIR detection.
\end{abstract}


\section{Introduction}\label{sec:intro}
Vision foundation models (VFMs) pretrained on massive RGB image collections have become strong general-purpose backbones for visual understanding \cite{clip, siglip, sam}. A central open question is whether the representations these models learn can extend beyond the modality they were pretrained on. The question is of practical consequence, as machine perception increasingly relies on sensors outside the visible spectrum. Near-infrared (NIR) \cite{nir-detection, nir_tech1}, short-wave infrared (SWIR) \cite{swir_advantage, rasmd2025}, and long-wave infrared (LWIR) \cite{mfnet, FMB} imaging enable perception in darkness, haze, and adverse weather across applications ranging from mobile robotics and driving to surveillance, yet in these bands the appearance and discriminative cues differ sharply from visible images.

An RGB-pretrained backbone can be highly capable on natural images yet brittle or inefficient when transferred to non-visible imagery. This phenomenon is commonly referred to as the spectral gap. Existing attempts to bridge this gap are narrow in scope. Work in remote sensing has produced multispectral and hyperspectral models for satellite imagery \cite{spectralgpt, s2mae, hypersigma}, and thermal-focused studies have leveraged the relatively abundant LWIR datasets available for ground-level tasks \cite{mspecgene, yuan2025unirgb}. However, no prior work has proposed a strategy for training a unified, generalist backbone capable of processing three infrared spectra alongside RGB, spanning NIR, SWIR, and LWIR, in a single model. Closing this gap is thus a concrete instance of a broader problem in modern AI, namely adapting large pretrained models to input distributions that lie far outside their pretraining data without retraining them from scratch.

The key challenge in spectral transfer of multiple infrared modalities is not just input adaptation but representation alignment. Each infrared wavelength produces responses to a scene that differ radically from visible light, even when sensors observe the same environment simultaneously. Therefore, to have a VFM capable of processing RGB, NIR, SWIR, and LWIR modalities, the distilled student model must learn to produce semantically consistent embeddings across visible and non-visible modalities without collapsing to trivial solutions or drifting from the teacher's semantic structure. Concretely, the same object (e.g., a pedestrian, a vehicle, a road marking) should elicit similar semantic activations regardless of whether it is observed in RGB, NIR, SWIR, or LWIR, even though its low-level appearance varies dramatically across sensors. 

To address this challenge, we introduce SpectraDINO, a parameter-efficient modality conditioned adaptation strategy that transfers a pretrained RGB VFM into a visible-infrared multispectral backbone without retraining from scratch. SpectraDINO updates a frozen ViT with lightweight, modality-specific components that address the spectral gap at two levels. At the input level, trainable spatial stems map single-channel infrared inputs into the backbone's representation space, while learnable modality embeddings condition all tokens on the active sensor type. At the feature level, per-modality bottleneck adapters inserted as residual modules into every transformer block enable modality-conditioned processing pathways without modifying any core attention weights. To align these adapted representations with the teacher's semantic structure, we employ a multi-objective distillation recipe spanning global CLS-token distillation, cross-modal contrastive learning, spatial patch-token correspondence, and a queue-based relational objective that preserves neighborhood structure in the embedding space. As naively optimizing these objectives jointly is unstable, we adopt a staged training protocol that progressively increases the number of trainable parameters and shifts the loss balance from strict feature mimicry toward broader cross-modal alignment. The result is a single, reusable backbone that achieves strong performance across RGB, NIR, SWIR, and LWIR on detection and segmentation benchmarks, without requiring modality-specific architectures or full task-specific finetuning. 

Our contributions are as follows:
\begin{itemize}[noitemsep]
    \item We propose SpectraDINO, a parameter-efficient spectral transfer framework that extends RGB vision foundation models to NIR, SWIR, and LWIR via modality-specific stems, token-level modality embeddings, and lightweight per-modality adapters without modifying the pretrained backbone's core weights.
    \item We introduce a multi-objective alignment recipe that transfers both pointwise and relational structure from the RGB teacher, combining cosine distillation, symmetric contrastive learning, patch-token alignment, and a queue-based neighborhood KL objective, trained under a staged protocol whose teacher-queue warmup and selective backbone fine-tuning prevent representation collapse.
    \item We demonstrate that the resulting backbone transfers effectively to dense perception, evaluating on multiple RGB-IR detection and segmentation benchmarks.
\end{itemize}

\section{Related Work}

\subsection{Vision Foundation Models}

Self-supervised pretraining on large-scale RGB data has become the dominant approach for producing transferable visual representations. Self-distillation and instance discrimination~\cite{dino, oquab2024dinov}, masked image modeling~\cite{bao2022beit, he2022mae}, and promptable segmentation~\cite{sam} have each demonstrated strong generalization across downstream tasks. However, these models are pretrained exclusively on visible imagery, limiting their direct applicability to infrared and multispectral sensors whose statistics differ substantially from those of RGB.

Contrastive vision-language pretraining popularized cross-modal alignment, later extended to arbitrary modality combinations~\cite{clip, jaegle2022perceiverio, girdhar2023imagebind}. Parameter-efficient methods such as visual prompts and lightweight adapters have meanwhile become standard for reusing pretrained ViTs without full fine-tuning, including for dense prediction~\cite{jia2022vpt, chen2022adaptformer, vitdet, chen2023vitadapter}, motivating our strategy of reusing strong RGB priors with minimal added capacity per sensing modality.

\subsection{Foundation Models for Multispectral Imaging}

In the remote sensing domain, several works have adapted self-supervised pretraining to satellite and hyperspectral imagery through spectral and temporal embeddings~\cite{s2mae, prithvieo20, spectralgpt, hypersigma}. These models target Earth observation data with many spectral bands and are not directly transferable to ground-level infrared perception.

Closer to our setting, InfMAE~\cite{infmae} investigates pretraining strategies tailored to infrared-specific statistics, PAD~\cite{pad} introduces patchwise-scale adapters for self-supervised infrared pretraining while preserving RGB representations, and UNIP~\cite{unip} analyzes how pretraining objectives shape spatial attention in the infrared domain and proposes hybrid-attention distillation. UniRGB-IR~\cite{yuan2025unirgb} injects cross-modal features into a frozen ViT backbone via lightweight adapter modules, IV-tuning~\cite{ivtuning} achieves parameter-efficient transfer through modality-aware prompts paired with rank-adaptive adapters, and M-SpecGene~\cite{mspecgene} proposes a generalized foundation model for RGB-thermal multispectral vision that learns shared representations across visible and thermal domains. In contrast to these works, which target a single infrared modality or a single modality pair, our goal is to bridge RGB visual foundation models to multiple multispectral sensing modalities simultaneously. 

\section{Methodology}\label{sec:method}

\subsection{SpectraDINO Architecture}
\label{sec:architecture}

\subsubsection{Teacher Backbone}
\label{sec:teacher}

SpectraDINO uses a frozen DINOv2~\cite{oquab2024dinov} model as the teacher. DINOv2 provides strong general-purpose visual features trained via self-supervised learning on a large curated RGB dataset. The teacher remains frozen throughout all training stages and processes the input without modification. It receives only RGB inputs and produces CLS token embeddings $\mathbf{z}_{\text{teacher}} \in \mathbb{R}^{D}$, where $D$ is the embedding dimension of the respective ViT variant ($D{=}768$ for ViT-B, $D{=}1024$ for ViT-L). These teacher embeddings serve as the semantic anchor for all alignment objectives described in Sec.~\ref{sec:alignment}. Freezing the teacher ensures that the target representation space remains stable throughout training, preventing drift in the distillation signal.

\begin{figure*}[t]
  \centering
  \includegraphics[width=\textwidth]{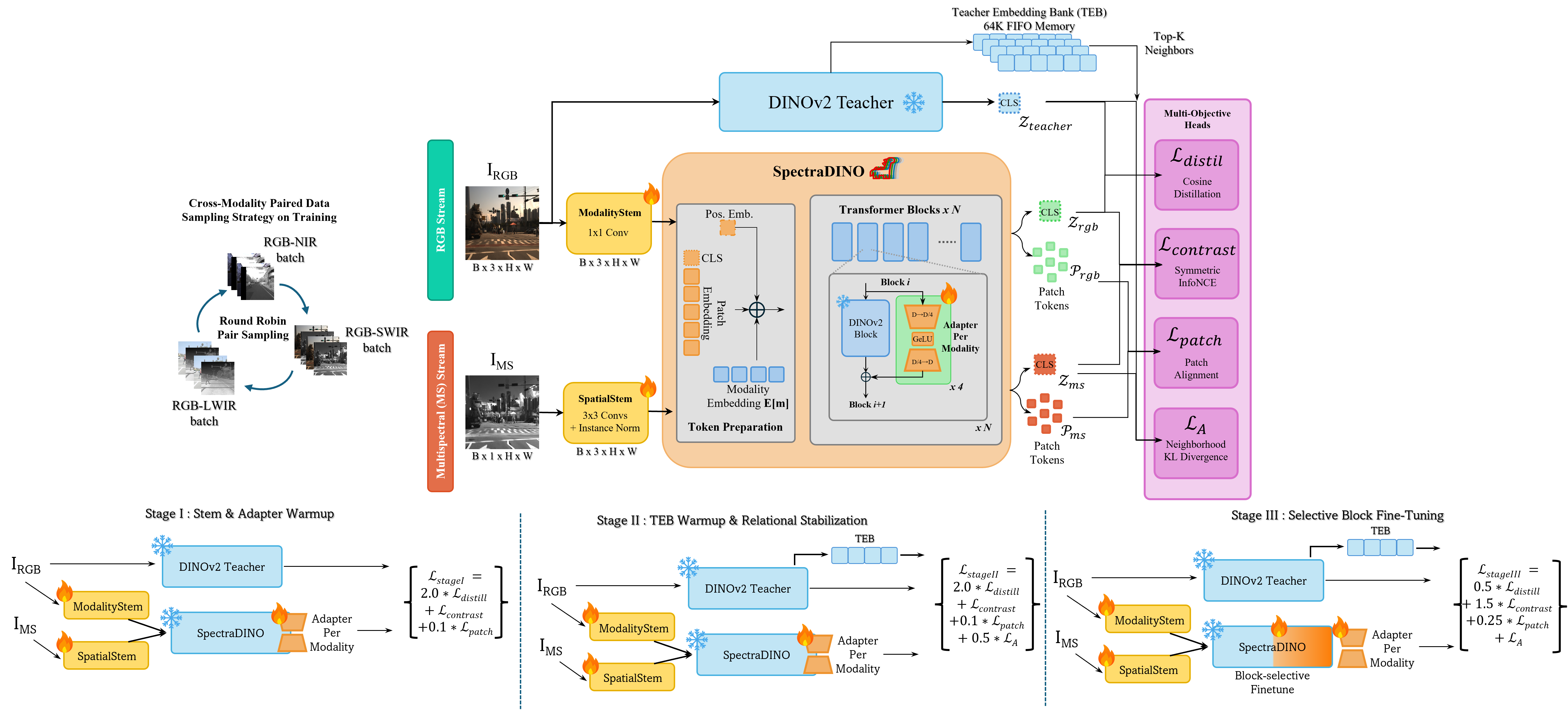}
  \caption{Overview of SpectraDINO. A frozen DINOv2 teacher provides target embeddings for RGB inputs. The student, SpectraDINO, processes both RGB and multispectral inputs through modality-specific stems into a shared ViT backbone augmented with per-block adapters and modality embeddings. All modalities share the same transformer weights, and only the stems, adapters, and modality embeddings are modality-conditioned. Training proceeds in three stages. Stage I warms up the stems and adapters with the backbone frozen, Stage II populates the teacher embedding bank (TEB) and activates the neighborhood loss, and Stage III unfreezes the deeper transformer blocks and rebalances the objectives toward cross-modal alignment.}
  \label{fig:architecture}
\end{figure*}

\subsubsection{Modality-Conditioned Input Stems}
\label{sec:stems}
Each spectral modality passes through a dedicated input stem that maps the raw input to a three-channel tensor compatible with the pretrained patch embedding layer of the ViT backbone.  For RGB inputs, we use a $1{\times}1$ convolution ($3 {\to} 3$ channels) initialized near identity. This design preserves the pretrained patch embedding behavior at initialization, ensuring that the student's RGB pathway starts close to the teacher's. The RGB stem remains frozen during Stage~I (Sec.~\ref{sec:stage1}) to maintain this alignment. For single-channel multispectral inputs (NIR, SWIR, LWIR), we use a spatial stem consisting of two convolutional layers:
{\small
\begin{equation}
  \text{SpatialStem}(\mathbf{x}) = \text{IN}\!\left(\text{Conv}_{3\times3}^{16 \to 3}\!\left(\sigma\!\left(\text{Conv}_{3\times3}^{1 \to 16}(\mathbf{x})\right)\right)\right),
  \label{eq:spatial_stem}
\end{equation}
}
where $\sigma$ denotes the GELU activation and IN denotes instance normalization. Instance normalization is applied at the output to handle the varying intensity distributions across spectral bands. Weights are initialized using Xavier uniform initialization. Each modality has its own stem instance with independent parameters.

\subsubsection{Modality-Conditioned Transformer Adapters}
\label{sec:adapters}
The core design principle of SpectraDINO is that all modalities share a single set of transformer weights. The pretrained ViT backbone is not duplicated or branched per modality. Instead, modality-specific behavior is achieved through two lightweight mechanisms: per-block bottleneck adapters and a learned modality embedding.

\paragraph{Per-block adapters.}
Each transformer block in the backbone is augmented with a bottleneck adapter applied as a residual connection after the block output:
{\small
\begin{equation}
  \mathbf{x}' = \text{Block}(\mathbf{x}) + \text{Adapter}_{m}(\text{Block}(\mathbf{x})),
  \label{eq:adapter}
\end{equation}
}
where $m \in \{0, 1, 2, 3\}$ indexes the modality (RGB, NIR, SWIR, LWIR). In practice, the block output $\text{Block}(\mathbf{x})$ is computed once and reused as input to the adapter, so the adapter introduces no redundant attention or feedforward computation. Each adapter is a two-layer MLP with a bottleneck of reduction factor 4. Weights are initialized with a small standard deviation ($\text{std}{=}0.01$) and biases are set to zero. This initialization ensures that adapter outputs are near-zero at the start of training, so the model begins as a close approximation of the original pretrained backbone and avoids destabilizing learned features.

Each transformer block maintains separate adapter parameters per modality. During a forward pass, samples in a batch are grouped by their modality index and routed to the corresponding adapter. For a ViT-B/14 backbone with 12 transformer blocks and 4 modalities, this yields 48 adapter instances, each containing $2 \times D \times D/4 = 2 \times 768 \times 192 = 294{,}912$ parameters. The total adapter overhead across all blocks and modalities is approximately 14.2M parameters for ViT-B/14, compared to 85.8M parameters in the frozen backbone. These adapters provide dedicated modality-conditioned capacity at a small parameter cost.

\paragraph{Modality embeddings.}  
A learned embedding table $\mathbf{E} \in \mathbb{R}^{4 \times D}$ maps each modality index to a $D$-dimensional vector. For a given input with modality index $m$, the corresponding embedding $\mathbf{E}[m]$ is added to every token (CLS and all patch tokens) after positional embeddings are applied:
{\small
\begin{equation}
  \mathbf{x}_{\ell=0} = \mathbf{x}_{\text{patch}} + \mathbf{e}_{\text{pos}} + \mathbf{E}[m],
  \label{eq:mod_emb}
\end{equation}
}
where $\mathbf{x}_{\text{patch}}$ denotes the patch-embedded input tokens, $\mathbf{e}_{\text{pos}}$ denotes positional embeddings, and $\mathbf{E}[m]$ is broadcast across the token dimension. This embedding provides a global conditioning signal that allows the shared self-attention layers to distinguish between modalities without requiring separate parameters in the backbone itself.

\paragraph{Shared backbone.}
The backbone transformer weights (self-attention projections, feedforward layers, layer norms within blocks) are identical for all modalities. Modality-specific computation is confined to the stems, adapters, and the embedding table. This weight sharing enforces modality-invariant intermediate representations and keeps the parameter count tractable when scaling to additional spectral bands.

\subsection{Multi-Objective Cross-Spectral Alignment}\label{sec:alignment}
Cross-spectral alignment requires supervision at multiple granularities, and each objective addresses a failure mode the others cannot. CLS-level distillation transfers global semantics but ignores spatial structure, contrastive learning enforces cross-modal pairing but does not anchor the student to the teacher, patch-level alignment preserves locality but imposes no manifold-level constraints, and the neighborhood objective preserves the relational geometry that pointwise losses leave unconstrained. SpectraDINO combines the four into
{\small
\begin{equation}
  \mathcal{L}_{\text{total}} = \lambda_d \, \mathcal{L}_{\text{distill}} + \lambda_c \, \mathcal{L}_{\text{contrast}} + \lambda_p \, \mathcal{L}_{\text{patch}} + \lambda_A \, \mathcal{L}_{A},
  \label{eq:total_loss}
\end{equation}
}
with stage-dependent weights $\lambda$ (see Table \ref{tab:loss_weights}). Table~\ref{tab:abl_lossfuncs} shows that adding each term improves downstream performance.

\subsubsection{Global distillation}\label{sec:distill}
The distillation loss aligns the CLS token embedding of the multispectral student and the frozen DINOv2 teacher using cosine distance. The teacher serves as a fixed semantic anchor. Minimizing $\mathcal{L}_{\text{distill}}$ transfers RGB-based semantics to the multispectral pathway without requiring labeled data. The frozen teacher also stabilizes training by providing a non-drifting target.

\subsubsection{Symmetric cross-modal contrastive learning}\label{sec:contrast}
$\mathcal{L}_{\text{contrast}}$ applies symmetric InfoNCE~\cite{oord2018representation} ($\tau{=}0.07$) between the student's RGB and multispectral embeddings, with paired images of the same scene as positives and all other in-batch combinations as negatives. Whereas distillation anchors the student to the teacher, this loss enforces consistency within the student across spectral bands.

\subsubsection{Patch-level spatial alignment}\label{sec:patch}
$\mathcal{L}_{\text{patch}}$ is the mean cosine distance between corresponding RGB and multispectral patch tokens over a random subset of $S$ of the $N$ patch positions, constraining the spatial feature arrangement that CLS-level objectives ignore. Subsampling keeps the cost low while covering all positions across iterations.

\subsubsection{Neighborhood structure preservation}\label{sec:neighborhood}
The pointwise objectives align individual pairs but not how samples relate to one another. $\mathcal{L}_{\text{A}}$ closes this gap by requiring the student to reproduce the teacher's local neighborhood topology. A Teacher Embedding Bank (TEB) $\mathbf{Q} \in \mathbb{R}^{K \times D}$ stores L2-normalized teacher CLS embeddings from recent batches with first-in-first-out updates after each optimizer step. For each sample $i$, we retrieve the top-$k$ entries of $\mathbf{Q}$ under teacher similarity and form, for both teacher and student, a similarity distribution over the retrieved neighbors,
{\small
\begin{equation}
  p_{\ast}^{(i,j)} = \frac{\exp\!\left(\mathbf{z}_{\ast}^{(i)} \cdot \mathbf{Q}_j / \tau\right)}{\sum_{l \in \text{top-}k} \exp\!\left(\mathbf{z}_{\ast}^{(i)} \cdot \mathbf{Q}_l / \tau\right)},
  \qquad \ast \in \{t, \text{ms}\},
  \label{eq:neighborhood_dist}
\end{equation}
}
and minimize $\mathcal{L}_{\text{A}} = \frac{1}{B}\sum_i \text{KL}(p_t^{(i)} \,\|\, p_{\text{ms}}^{(i)})$. The bank holds $K{=}65{,}536$ entries, roughly a quarter of our pretraining pairs, and is checkpointed across stages. Since it is a sliding snapshot of the data distribution rather than a persistent memory, FIFO eviction discards nothing the loss depends on, and the frozen teacher ensures stored embeddings never go stale.

\subsection{Training Strategy}\label{sec:training_strategy}
We train SpectraDINO in three stages, each of which resumes from the previous checkpoint. The staging serves two purposes. First, it prevents newly initialized modality-specific parameters from destabilizing the pretrained backbone early in training. Second, it lets objectives that depend on accumulated state, namely the TEB for $\mathcal{L}_{\text{A}}$, warm up before influencing gradients. All stages use AdamW~\cite{loshchilov2017decoupled}, mixed precision, and a warmup-cosine schedule. Each step processes a batch of paired RGB and multispectral images from a single modality, with modalities sampled in round-robin order so each receives equal training signal per epoch.

Loss weights and hyperparameters are identical across all backbone variants, and variant-specific tuning is limited to learning rate, batch size, epoch count, and unfreeze depth (supp. Table A1). Table~\ref{tab:loss_weights} summarizes the stage configuration.

\subsubsection{Stage I: Stem and Adapter Warmup}\label{sec:stage1}

In Stage~I, the pretrained ViT backbone is entirely frozen. Only the multispectral stems, per-block adapters, modality embeddings, and the final layer norm are trainable. The purpose of Stage~I is to learn modality routing, which means mapping raw NIR, SWIR, and LWIR inputs into the pretrained RGB feature space through the stems and adapters, without modifying the backbone representations. The high distillation weight ensures that multispectral embeddings are pulled close to the teacher. The lower contrastive weight provides a secondary cross-modal alignment signal. Patch alignment operates on a small subset of tokens (25\%) to regularize spatial features at low cost.

\subsubsection{Stage II: Queue Warmup and Relational Stabilization}\label{sec:stage2}
Stage~II is a short bridge stage that introduces $\mathcal{L}_{A}$ while the backbone remains frozen. Training resumes from the Stage~I best checkpoint with no warmup. $\mathcal{L}_{A}$ has its own internal warmup, with $\lambda_A$ set to 0 for the first epoch, during which the TEB accumulates embeddings without producing gradients. Starting from the second epoch, TEB contains representative teacher embeddings and $\mathcal{L}_{A}$ begins contributing to the loss. Stage~II exists because $\mathcal{L}_{A}$ requires a populated queue to produce meaningful gradients. Activating $\mathcal{L}_{A}$ simultaneously with backbone unfreezing (Stage~III) would introduce instability due to a new loss signal and newly trainable backbone parameters. The bridge stage decouples these by stabilizing $\mathcal{L}_{A}$ first.

\subsubsection{Stage III: Selective Backbone Fine-Tuning}\label{sec:stage3}

Stage~III unfreezes the deeper transformer blocks while keeping the earlier blocks frozen, and the TEB state carries over. Loss weights are rebalanced. $\lambda_d$ drops to 0.5 since strict alignment to the RGB teacher becomes less appropriate as the backbone adapts to non-visible spectra, $\lambda_c$ rises to 1.5 to maintain cross-modal consistency while backbone weights move, $\lambda_p$ rises to 0.25 with a 0.5 sampling ratio for stronger spatial supervision, and $\mathcal{L}_{\text{A}}$ runs at full weight to preserve the teacher's neighborhood structure during fine-tuning. The final model therefore combines pretrained adapters with partial backbone adaptation, and neither ingredient suffices alone (Sec. \ref{sec:ablation}).

\begin{table}[t]
    \centering
    \small
    \caption{Loss weights and $\mathcal{L}_{A}$ configuration across training stages. These values are shared across all backbone variants. "---" indicates the loss is disabled. PSR: patch sample ratio, K: Queue size}
    \label{tab:loss_weights}
    \begin{tabular}{lccc}
      \toprule
      & St.~I & St.~II & St.~III \\
      \midrule
      Backbone & Frozen & Frozen & Partial \\
      \midrule
      $\lambda_d$ ($\mathcal{L}_{\text{dist}}$) & 2.0 & 2.0 & 0.5 \\
      $\lambda_c$ ($\mathcal{L}_{\text{con}}$) & 1.0 & 1.0 & 1.5 \\
      $\lambda_p$ ($\mathcal{L}_{\text{ptch}}$) & 0.1 & 0.1 & 0.25 \\
      PSR & 0.25 & 0.25 & 0.5 \\
      $\lambda_A$ ($\mathcal{L}_{A}$) & --- & 0.5 & 1.0 \\
      $K$ & --- & 65{,}536 & 65{,}536 \\
      Top-$k$ & --- & 128 & 128 \\
      $\tau$ & --- & 0.07 & 0.07 \\
      \bottomrule
    \end{tabular}
\end{table}

\section{Experiments}\label{sec:experiments}

\subsection{Pretraining Setup}\label{sec:pretraining_setup}
We pretrain SpectraDINO on 300K paired cross-spectral images, comprising 100K RGB--NIR pairs from MS2~\cite{kaist_ms2}, 100K RGB--SWIR pairs from RASMD~\cite{rasmd2025}, and 100K RGB--LWIR pairs from IR500K ~\cite{ir500k}, sampled in round-robin order so that no spectral domain dominates optimization. We train ViT-B/14 and ViT-L/14 variants initialized from the corresponding DINOv2 checkpoints, following the three-stage procedure of Sec.~\ref{sec:training_strategy} with 100, 10, and 75 epochs per stage. Stage~III unfreezes half of the transformer blocks (6 of 12 for ViT-B, 12 of 24 for ViT-L). Per-variant learning rates, batch sizes, and remaining implementation details are listed in the supplementary material. All pretraining runs use two NVIDIA A6000 GPUs.

\subsection{Downstream Evaluation Setup}\label{sec:downstream_setup}
We evaluate on object detection and semantic segmentation across seven benchmarks covering three spectral modalities: LWIR (FLIR~\cite{flir}, LLVIP~\cite{llvip}, MFNet~\cite{mfnet}, SemanticRT~\cite{ecm}, FMB~\cite{FMB}), SWIR (RASMD), and NIR (MCubeS~\cite{mcubes}). To our knowledge, no annotated SWIR segmentation and NIR detection benchmarks are publicly available, so SWIR is evaluated on detection and NIR is evaluated on segmentation only. We report COCO-style mAP and AP$_{50}$ for detection and mIoU with mean class accuracy (mAcc) for segmentation. The same fine-tuning recipe is applied uniformly across all benchmarks, training the task head, adapters, and stems without per-dataset tuning. Unless noted otherwise, baseline results are taken from the original publications, and retrained baselines are trained by us under matched protocols.

\subsubsection{Feature Extraction and Fusion.}\label{sec:dual_stream}

Each input's RGB and MS components are processed sequentially through the shared backbone, producing two token sequences of dimension $D$ that are concatenated and projected back to $D$ dimensions by a linear layer with LayerNorm and GELU. Fusion is applied at the final block for detection with ViTDet and at multiple intermediate blocks for multi-scale segmentation heads. We deliberately use this parameter-light fusion to isolate the contribution of the pretrained representations from fusion-module design, and show in the supplementary that stronger fusion (cross-attention) yields further gains (supp. Table A2).

\subsection{Object Detection Experiments}\label{sec:detection}

\begin{table}
\small
\centering
\setlength{\tabcolsep}{2.5pt}
\caption{Quantitative comparison on the FLIR and LLVIP datasets. The best and second best results are highlighted in \textbf{bold} and \underline{underline}, respectively.}
\begin{tabular}{lcccc}
\toprule
\multirow{2}{*}{Model}
 & \multicolumn{2}{c}{FLIR (LWIR)} 
 & \multicolumn{2}{c}{LLVIP (LWIR)} \\
\cmidrule(lr){2-3}\cmidrule(lr){4-5}
 & mAP & AP$_{50}$  
 & mAP & AP$_{50}$ \\
\midrule
GAFF \cite{zhang2021guided} & 37.4 & 74.7 & 55.8 & 94.0 \\
CSAA\cite{cao2023multimodal} & 41.3 & 79.2 & 59.2 & 94.3 \\
TIRDet \cite{wang2023tirdet} & \underline{44.3} & 81.4 & 64.2 & 96.3 \\
MMI-Det \cite{zeng2024mmi} & 40.5 & 79.8 & 64.4 & \textbf{98.9} \\
CrossFormer \cite{lee2024crossformer} & 42.1 & 79.3 & 65.1 & 97.4 \\
UniRGB-IR \cite{yuan2025unirgb} & 44.1 & 81.4 & 63.2 & 96.1 \\
M-SpecGene \cite{mspecgene} & \textbf{44.7} & \textbf{84.8} & \underline{65.3} & 97.4 \\
\textbf{SpectraDINO}$_{Base}$ & 40.7 & 78.2 & 63.5 & 97.6 \\
\textbf{SpectraDINO}$_{Large}$ & \underline{44.3} & \underline{83.6} & \textbf{65.7} & \underline{98.1} \\

\bottomrule
\end{tabular}
\label{tab:flir_llvip}
\end{table}

We evaluate detection performance using the ViTDet head. As loss functions, we use focal loss for classification and GIoU loss for box regression. We train with AdamW using per-module learning rates: $1 \times 10^{-5}$ for unfrozen backbone blocks, $2.5 \times 10^{-5}$ for adapters and stems, and $1 \times 10^{-4}$ for the detection head. We evaluated different training strategies, including only training the detection head, training the detection head and adapters, half-unfreezing the backbone, and full-unfreezing the backbone. For most datasets, finetuning the adapters, the head, and half of the backbone achieved the best results. The second most successful strategy was fine-tuning the adapters, stems, and the detection head. An ablation study of these training strategies is given in Table \ref{tab:dinov2_baselines}. Tables~\ref {tab:flir_llvip} and~\ref {tab:rasmd} summarize the detection results across three benchmarks spanning two spectral bands.

\begin{table}
\small
\centering
\caption{Quantitative comparison on RASMD (SWIR) dataset. The best and second best results are highlighted in \textbf{bold} and \underline{underline}, respectively. * denotes baselines retrained by us on RASMD under our matched protocol.}
\begin{tabular}{lcc}
\toprule
Model & mAP & AP$_{50}$ \\
\midrule
MRFA* ~\cite{MRFA} & 21.4 & 30.0 \\
RGBX-Fusion* ~\cite{rgbxfusion} & 26.1 & 37.0 \\
UniRGB-IR* \cite{yuan2025unirgb} & \underline{46.6} & 68.3 \\
M-SpecGene* \cite{mspecgene} & 45.1 & 68.8 \\
\textbf{SpectraDINO}$_{Base}$ & 45.8 & \underline{73.1} \\
\textbf{SpectraDINO}$_{Large}$ & \textbf{49.3} & \textbf{77.2} \\
\bottomrule
\end{tabular}
\label{tab:rasmd}
\end{table}

On the LWIR benchmarks, SpectraDINO$_\text{Large}$ ties TIRDet for the second-best mAP on FLIR (44.3) and records the second-best AP$_{50}$ (83.6), trailing only M-SpecGene. On LLVIP it achieves the best mAP of 65.7, ahead of M-SpecGene, while its AP$_{50}$ of 98.1 is second to MMI-Det. We attribute the remaining gaps on the LWIR benchmarks largely to the design focus of the strongest competitors. Detectors such as TIRDet and MMI-Det are purpose-built for RGB-thermal fusion and incorporate mechanisms tailored to extract maximal complementary information from this specific modality pair, whereas we design the SpectraDINO backbone with lightweight modality stems and adapters that is not specialized for any one band. The SWIR results on RASMD (Table~\ref{tab:rasmd}) present the strongest case for spectral generalization. SpectraDINO$_\text{Large}$ reaches 49.3 mAP and 77.2 AP$_{50}$, surpassing the strongest finetuned baselines by 2.7 mAP over UniRGB-IR and 8.4 AP$_{50}$ over M-SpecGene. SpectraDINO$_\text{Base}$ remains competitive at 45.8 mAP with the second-best AP$_{50}$ of 73.1, without any per-dataset architectural changes. This pattern is consistent with SpectraDINO's design. On LWIR, where abundant paired data has enabled years of architecture specialization, a band-agnostic backbone remains on par with the best dedicated models. On SWIR, for which those specialized pipelines were never optimized, the shared spectral representation learned during pretraining transfers
directly, and the widening margin from Base to Large suggests this benefit grows with backbone capacity.

\subsection{Semantic Segmentation Experiments}\label{sec:segmentation}
For semantic segmentation, we adopt the UPerNet~\cite{xiao2018unified} head, a widely used and well-established choice for evaluating vision foundation models on dense prediction. The head is trained with per-pixel cross-entropy loss. Tables~\ref{tab:mfnet}, ~\ref{tab:semanticrt}, and ~\ref{tab:mcubes} report segmentation results on four benchmarks spanning the LWIR and NIR bands.

Across the three LWIR benchmarks, SpectraDINO$_\text{Large}$ achieves the best mIoU on SemanticRT and FMB, exceeding the strongest baselines by 1.4 and 2.6 points, respectively. On MFNet it obtains the best mAcc of 77.4 while its mIoU of 60.8 trails PEAFusion(L) by 1.5 points and HAPNet by 0.7. As in the detection experiments, the methods ahead of SpectraDINO on this benchmark are dedicated RGB-thermal architectures whose fusion mechanisms are designed to extract maximal complementary information from this particular modality pair, while SpectraDINO applies the same band-agnostic backbone across all experiments. The combination of leading mAcc with slightly lower mIoU also suggests that the model identifies the correct classes reliably but loses some accuracy along fine object boundaries, a plausible consequence of patch-level ViT features on the low-resolution MFNet imagery. Scaling from SpectraDINO$_\text{Base}$ to SpectraDINO$_\text{Large}$ improves mIoU by 4.1, 7.7, and 1.9 points on SemanticRT, FMB, and MFNet, respectively, indicating that the pretrained representations continue to benefit from additional backbone capacity. For NIR segmentation, we evaluate on MCubeS (Table~\ref{tab:mcubes}), a material segmentation benchmark providing RGB, NIR, angle of linear polarization, and degree of linear polarization. All compared methods use the full four-modality input, whereas SpectraDINO operates on RGB and NIR alone. Despite this disadvantage, SpectraDINO$_\text{Large}$ reaches 46.76 mIoU, surpassing four of the six four-modality methods. The remaining gap of 4.8 points to CMNeXt suggests that polarization carries complementary cues for material boundaries that RGB-NIR features cannot fully recover. Still, the pretrained NIR representations close a large part of the missing-modality gap while using half the input modalities.

\begin{table}[t]
\small
\centering
\caption{Quantitative comparison on the MFNet (LWIR) and FMB (LWIR) datasets. The best and second best results are highlighted in \textbf{bold} and \underline{underline}, respectively.}
\label{tab:mfnet}
\begin{tabular}{clcc}
\toprule
Dataset & Model & mAcc & mIoU \\
\midrule
\multirow{7}{*}{\textit{MFNet}} & MFNet~\cite{mfnet}   & 45.1 & 39.7 \\
& RTFNet~\cite{rtfnet} & 63.1 & 53.2 \\
& EGFNet~\cite{egfnet} & 72.7 & 54.8 \\
& HAPNet~\cite{hapnet} & 70.3 & \underline{61.5} \\
& PEAFusion(B) ~\cite{peafusion} & \underline{77.0} & 61.0 \\
& PEAFusion(L) ~\cite{peafusion} & 72.6 & \textbf{62.3} \\
& \textbf{SpectraDINO}$_{Base}$    & 75.2 & 58.9 \\
& \textbf{SpectraDINO}$_{Large}$   & \textbf{77.4} & 60.8 \\
\midrule
\multirow{8}{*}{\textit{FMB}} & YDTR~\cite{yd_tr} & 64.5 & 56.5 \\   
& SwinFusion~\cite{swin_fusion} & 64.7 & 56.4 \\   
& ReCoNet~\cite{reconet} & 64.5 & 56.4 \\
& MoEFusion~\cite{moefusion} & 64.5 & 57.0 \\ 
& SuperFusion~\cite{superfusion} & 64.0 & 55.9 \\
& PEAFusion~\cite{peafusion} & \textbf{85.3} & \underline{69.8} \\
& \textbf{SpectraDINO}$_{Base}$ & 72.7 & 64.7 \\
& \textbf{SpectraDINO}$_{Large}$ & \underline{80.3} & \textbf{72.4} \\
\bottomrule
\end{tabular}
\end{table}

\begin{table}[t]
\small
\centering
\setlength{\tabcolsep}{2.5pt}
\caption{Quantitative comparison on the SemanticRT dataset. The best and second best results are highlighted in \textbf{bold} and \underline{underline}, respectively.}
\begin{tabular}{lc}
\toprule
\multirow{2}{*}{Model} & SemanticRT (LWIR) \\
 & mIoU \\
\midrule
MFNet \cite{mfnet} & 74.08 \\
RTFNet \cite{rtfnet} & 75.48 \\
EGFNet \cite{egfnet} & 77.44 \\
ECM \cite{ecm} & 79.26 \\
UniRGB-IR \cite{yuan2025unirgb} & 75.21 \\
M-SpecGene \cite{mspecgene} & \underline{79.84} \\
\textbf{SpectraDINO}$_{Base}$ & 77.17 \\
\textbf{SpectraDINO}$_{Large}$ & \textbf{81.24} \\
\bottomrule
\end{tabular}
\label{tab:semanticrt}
\end{table}

\begin{table}[t]
\small
\centering
\caption{Quantitative comparison on the MCubeS (NIR) dataset. R: RGB, A: angle of lin. polarization, D: degree of lin. polarization, N: NIR, $\dagger$: results with only two modalities.}
\label{tab:mcubes}
\begin{tabular}{lccl}
\toprule
\multirow{2}{*}{Model}
 & \multicolumn{3}{c}{MCubeS (NIR)} \\
\cmidrule(lr){2-4}
 & Modalities & & mIoU \\
\midrule
DeepLabv3+~\cite{chen2018encoder} & R-A-D-N & & 38.13 \\
MMTM~\cite{joze2020mmtm} & R-A-D-N & & 39.71 \\
FuseNet~\cite{fusenet} & R-A-D-N & & 40.58 \\
MCubeSNet~\cite{mcubes} & R-A-D-N & & 42.46 \\
CBAM~\cite{woo2018cbam} & R-A-D-N & & \underline{51.32} \\
CMNeXt~\cite{zhang2023delivering} & R-A-D-N & & \textbf{51.54} \\
\midrule
\textbf{SpectraDINO}$_{Base}$ & \underline{R-N}$\dagger$ & & 41.89$\dagger$ \\
\textbf{SpectraDINO}$_{Large}$ & \underline{R-N}$\dagger$ & & 46.76$\dagger$ \\
\bottomrule
\end{tabular}
\end{table}

\subsection{Ablation Study}\label{sec:ablation}

\begin{table}
\centering
\small
\setlength{\tabcolsep}{3.5pt}
\caption{Ablation on loss function combinations during Stage III of representation learning with SpectraDINO\textsubscript{Base}. We report semantic segmentation on FMB and object detection on RASMD.}
\label{tab:abl_lossfuncs}
\begin{tabular}{cccc cc ccc}
\toprule
\multicolumn{4}{c}{Loss terms} & \multicolumn{2}{c}{Seg. (FMB)} & \multicolumn{3}{c}{Det. (RASMD)} \\
\cmidrule(lr){1-4} \cmidrule(lr){5-6} \cmidrule(lr){7-9}
$\mathcal{L}_{\text{dis}}$ & $\mathcal{L}_{\text{con}}$ & $\mathcal{L}_{\text{ptch}}$ & $\mathcal{L}_{A}$ & mAcc & mIoU & AP$_{50}$ & AP$_{75}$ & mAP \\
\midrule
\cmark & & & & 65.25 & 58.91 & 64.73 & 34.21 & 37.61 \\
\cmark & \cmark & & & 67.34 & 60.83 & 69.27 & 38.49 & 37.68 \\
\cmark & \cmark & \cmark & & 71.83 & 63.70 & 70.89 & 39.89 & 40.17 \\
\cmark & \cmark & \cmark & \cmark & \textbf{72.71} & \textbf{64.72} & \textbf{73.12} & \textbf{44.43} & \textbf{43.24} \\
\bottomrule
\end{tabular}
\end{table}

We ablate the Stage III loss composition, the contribution of the bottleneck adapters, and the finetuning strategy, all on SpectraDINO\textsubscript{Base}. Additional ablations on fusion strategy and training stage progression selection are provided in Appendices A2 and A3, respectively.

\subsubsection{Are all four training objectives necessary?} Table~\ref{tab:abl_lossfuncs} adds the Stage III loss terms
cumulatively, evaluated on both an LWIR segmentation benchmark (FMB) and a SWIR detection benchmark (RASMD) so that conclusions do not depend on a single dataset or band. Every term contributes, and the full combination improves over the distillation-only baseline by 5.8 mIoU and 5.6 mAP. The two added terms play complementary roles. $\mathcal{L}_{\text{ptch}}$ brings the largest segmentation gain (+2.9 mIoU), while $\mathcal{L}_{\text{A}}$ contributes most to detection (+3.1 mAP) with the gain concentrated at the strict threshold, which indicates that preserving the teacher's neighborhood structure primarily sharpens instance-level localization. Beyond its direct contribution, we found $\mathcal{L}_{\text{A}}$ necessary for stable optimization, as runs without it were prone to divergence of the contrastive objective during backbone unfreezing.

\begin{table}[t]
\centering
\small
\caption{Effect of module finetuning. All models are Base-size. The top block has no cross-spectral pretraining but the SpectraDINO includes it. All rows share input format, fusion, task heads, and evaluation protocol; a single representative seed is reported. Repeated runs of the selectively tuned baseline varied by less than 0.5 mAP and 0.1 mIoU across seeds.}
\setlength{\tabcolsep}{2.3pt}
\begin{tabular}{lcrrrr}
\toprule
 & & \#Par. & RASMD & MFNet & Sem.RT \\
Model & Tuning & (M) & mAP & mIoU & mIoU \\
\midrule
DINOv2 & frozen & 8.8 & 37.3 & 57.0 & 72.4 \\
DINOv2 & adapt.(rand.) & 23.0 & 30.6 & 31.8 & 68.2 \\
DINOv2 & top 6/12 & 52.3 & 43.2 & 58.4 & 73.0 \\
DINOv2 & full & 95.5 & 45.0 & 47.6 & 72.0 \\
\midrule
\textbf{SpectraDINO} & adapt. \& stems & 23.3 & 39.3 & 55.8 & 72.2 \\
\textbf{SpectraDINO} & top 6/12 & 65.8 & 45.8 & \textbf{58.9} & \textbf{77.2} \\
\textbf{SpectraDINO} & full & 99.8 & 41.9 & 51.4 & 70.7 \\
\bottomrule
\end{tabular}
\label{tab:dinov2_baselines}
\end{table}

\subsubsection{Do the gains come from the adapter architecture, from partial finetuning, or from cross-spectral pretraining?}
Table~\ref{tab:dinov2_baselines} separates the three factors by comparing against DINOv2 baselines of identical backbone size that share the input format, fusion, task heads, tuning regime, and evaluation protocol. The adapter architecture alone explains none of the improvement. Attaching randomly initialized adapters to a frozen DINOv2 degrades performance sharply (30.6 vs.\ 37.3 mAP on RASMD) despite nearly tripling the trainable parameter count, ruling out added capacity as the source of gains. Cross-spectral pretraining is what makes the added parameters useful. At the matched top-6/12 tuning regime, SpectraDINO surpasses the identically tuned DINOv2 by 2.6 mAP on RASMD and 4.1 mIoU on SemanticRT. With a fully frozen backbone, the pretrained adapters improve SWIR detection (+2.0 mAP) but not LWIR segmentation, so we do not claim that adapters alone suffice; the pretrained spectral knowledge is realized fully only when the upper backbone blocks can co-adapt. Finally, unfreezing all blocks reverses the trend, dropping 3.9 mAP on RASMD relative to the half-unfrozen setting despite training 34M more parameters, which we attribute to the small downstream datasets eroding the pretrained representations. The half-unfrozen configuration balances adaptation against preservation, and we adopt it as the default for all reported results.

\section{Conclusion}\label{sec:conclusion}
We presented SpectraDINO, which adapts an RGB vision foundation model to NIR, SWIR, and LWIR imagery via modality-conditioned stems and per-block bottleneck adapters, and a three-stage training pipeline that progresses from a frozen adapter warmup to partial backbone finetuning under a unified objective that combines distillation, contrastive alignment, patch-level matching, and neighborhood consistency. The gains are largest where prior coverage is thinnest, reaching 49.3 mAP on SWIR detection, 2.7 points above the strongest finetuned baseline, while remaining competitive with specialized thermal methods on LWIR and outperforming several polarization-assisted baselines on NIR material segmentation with RGB-NIR input alone. Our ablations show that these gains stem from cross-spectral pretraining rather than from the adapter architecture or the tuning budget, and that they are realized fully only when the upper backbone blocks co-adapt. The residual gap to specialized methods on LWIR benchmarks deserves comment. Unlike NIR and SWIR, which capture reflected light and therefore share illumination and material structure with RGB, LWIR imagery is dominated by emitted thermal radiation, so object appearance largely decouples from the reflectance cues the RGB teacher encodes. We hypothesize that this makes LWIR the hardest band to align within a single shared representation, and that dedicated RGB-thermal architectures retain an edge precisely because their fusion mechanisms are engineered around this asymmetry. Quantifying cross-band alignment and treating emissive and reflective bands asymmetrically during pretraining are promising directions for further exploration.

\section*{Appendix}

\setcounter{table}{0}
\renewcommand{\thetable}{A\arabic{table}} 

\subsection*{A1. SpectraDINO variant-specific training parameters}

Table~\ref{tab:variant_params} lists the training parameters used in the representation learning phase of SpectraDINO for both backbone variants. Learning rate and per-GPU batch size follow model capacity, while the stage schedule of 100, 10, and 75 epochs is shared across variants. Stage~III unfreezes the top half of the backbone in both cases, so the number of trainable blocks scales with depth rather than with a fixed budget.

\begin{table}[ht!]
    \small
    \centering
    \caption{Variant-specific training parameters. Learning rate and batch size scale with model capacity. *$D$: Embedding Dimension, AB: Adapter Bottleneck, UnfB: Unfrozen Blocks.}
    \label{tab:variant_params}
    \setlength{\tabcolsep}{1.9pt}
    \begin{tabular}{lcc}
      \toprule
      & ViT-B/14 & ViT-L/14 \\
      \midrule
      $D$* & 768 & 1024 \\
      AB* & 192 & 256 \\
      Total blocks & 12 & 24 \\
      \midrule
      \multicolumn{2}{l}{\textit{Stage I}} \\
      \quad Epochs & 100 & 100 \\
      \quad LR & $1{\times}10^{-4}$ & $8{\times}10^{-5}$ \\
      \quad BS/GPU & 128 & 64 \\
      \midrule
      \multicolumn{2}{l}{\textit{Stage II}} \\
      \quad Epochs & \multicolumn{2}{c}{10} \\
      \quad LR & $1{\times}10^{-4}$ & $8{\times}10^{-5}$ \\
      \quad BS/GPU & 128 & 64 \\
      \midrule
      \multicolumn{2}{l}{\textit{Stage III}} \\
      \quad Epochs & 75 & 75 \\
      \quad LR & $4{\times}10^{-5}$ & $3{\times}10^{-5}$ \\
      \quad BS/GPU & 128 & 64 \\
      \quad UnfB* & 6 (50\%) & 12 (50\%) \\
      \bottomrule
    \end{tabular}
  \end{table}

\subsection*{A2. Fusion Strategy Comparison}

All downstream experiments in the main paper use feature concatenation as the fusion strategy. Concatenation adds no learnable parameters, so any gain is attributable to the pretrained representations rather than to a fusion module. Here, we test how the same representations behave under more expressive fusion.

\begin{table}[ht!]
\centering
\caption{Ablation on the effect of feature fusion strategies on downstream object detection performance. Adapters and modality stems of SpectraDINO\textsubscript{Base} are finetuned on RASMD with concatenation, cross-attention, and gated-token fusion.}
\label{tab:abl_fusion}
\begin{tabular}{lccc}
\toprule
Fusion strategy & mAP$_{50}$ & mAP$_{75}$ & mAP \\
\midrule
Baseline (Concat) & 73.12 & 47.83 & 45.81 \\
Cross attention & 74.01 & 48.59 & 46.34 \\
Gated token & 71.21 & 39.98 & 39.08 \\
\bottomrule
\end{tabular}
\end{table}
 
Table~\ref{tab:abl_fusion} reports RASMD detection results when concatenation is replaced by two alternative strategies with all other settings fixed. Cross-attention fusion, which computes pairwise attention between modality-specific feature maps, improves all three metrics, raising mAP$_{50}$ from 73.12 to 74.01 (+0.89), mAP$_{75}$ from 47.83 to 48.59 (+0.76) and mAP from 45.81 to 46.34 (+0.53). Gated-token fusion, which learns a per-token weight to blend modality features, falls behind the baseline on every metric and reaches 71.21 mAP$_{50}$ and 39.08 mAP. We attribute this to data scale, since a per-token gating policy has to be fit from scratch and RASMD is small relative to the pretraining corpus.

Two points follow. First, SpectraDINO representations are compatible with more expressive fusion and gain from it, although the margin from cross-attention is under 1 point across all metrics and comes at the cost of additional downstream parameters. Second, plain concatenation is already competitive, as expected if the modalities are aligned during pretraining rather than at the fusion stage. We keep concatenation throughout the main paper for this reason, and we encourage future work to explore task-specific fusion architectures built on SpectraDINO features at larger downstream data scales.

\subsection*{A3. Effect of Representation Training Stages}

\begin{table}[ht!]
    \centering
    \caption{Ablation on the effect of each training stage on downstream task performance. Representations are evaluated after Stage I, II, and III using SpectraDINO\textsubscript{Base}.}
    \label{tab:abl_stages}
    
    \begin{tabular}{lcc}
    \toprule
    Configuration & mAcc & mIoU \\
    \midrule
    Stage I & 67.61 & 60.84 \\
    Stage II (Bridge) & 66.79 & 66.79 \\
    Stage III & 72.71 & 64.72 \\
    \bottomrule
    \end{tabular}
    
    \vspace{2mm}
    {\footnotesize Results of semantic segmentation task on FMB Dataset\par}
    
    \vspace{4mm} 
    
    \begin{tabular}{lccc}
    \toprule
    Configuration & AP$_{50}$ & AP$_{75}$ & mAP \\
    \midrule
    Stage I & 61.04 & 34.51 & 35.09 \\
    Stage II (Bridge) & 63.16 & 33.83 & 35.09 \\
    Stage III & 73.12 & 47.83 & 45.81 \\
    \bottomrule
    \end{tabular}
    
    \vspace{2mm}
    {\footnotesize Results of object detection task on RASMD Dataset\par}

\end{table}

We evaluate the learned representations after each training stage to quantify the contribution of the staged curriculum. Table~\ref{tab:abl_stages} reports downstream results on FMB and RASMD using SpectraDINO\textsubscript{Base}.

After Stage~I the model reaches 67.61 mAcc and 60.84 mIoU on FMB, and 61.04 AP$_{50}$ with 35.09 mAP on RASMD. The backbone still holds its original DINOv2 RGB weights at this point, and only the modality stems and adapters have been trained. Lightweight modules alone are therefore enough to route non-RGB input into a frozen RGB representation and obtain usable multispectral features.

Stage~II introduces the neighborhood KL divergence loss $\mathcal{L}_A$ with the FIFO teacher embedding bank. Its effect is a reorganization of the embedding neighborhood rather than an across-the-board gain. On FMB, mIoU rises from 60.84 to 66.79 (+5.95) while mAcc falls slightly from 67.61 to 66.79. On RASMD, AP$_{50}$ rises from 61.04 to 63.16 (+2.12), while AP$_{75}$ moves from 34.51 to 33.83 and mAP is unchanged at 35.09. This is consistent with the stage's role. It runs for only 10 epochs and is intended to calibrate the representation space before backbone unfreezing, not to add discriminative capacity.

Stage~III, which unfreezes the top half of the backbone under the rebalanced full-loss ensemble, produces the largest gains. On RASMD, every metric improves, with AP$_{50}$ reaching 73.12 (+9.96 over Stage~II), AP$_{75}$ reaching 47.83 (+14.00), and mAP reaching 45.81 (+10.72). On FMB the picture is mixed. mAcc reaches 72.71 (+5.92 over Stage~II) while mIoU settles at 64.72, which is 2.07 points below the Stage~II value and 3.88 above Stage~I. Co-adapting the upper transformer blocks with the stems and adapters improves per-class recall at the cost of boundary precision, an effect visible only on the denser segmentation task.

Taken together, the three stages contribute in different ways. Stage~I fits the lightweight modules against a fixed representation, Stage~II calibrates the embedding neighborhood, and Stage~III refines the backbone once both are stable. The ordering matters because unfreezing the backbone before the stems and adapters have converged exposes the pretrained RGB representation to gradients from randomly initialized modules, which is the setting in which forgetting is most likely. Detection benefits from the full curriculum across every metric, while segmentation trades a small drop in mIoU for a larger gain in mAcc.

\subsection*{Discussion on Limitations}

\paragraph{Absence of cross-spectral paired data.}
Our pretraining data consists only of bimodal pairs (RGB–NIR, RGB–SWIR, and RGB–LWIR). As a result, the model never observes simultaneous relationships between non-RGB modalities. During training, RGB therefore acts as an intermediary alignment anchor. NIR, SWIR, and LWIR are each aligned to RGB, but no direct alignment is enforced between infrared modalities. This design implicitly assumes that the RGB embedding space provides a sufficiently structured reference for all infrared bands. While this assumption is reasonable for spectrally adjacent modalities such as NIR and SWIR, it becomes weaker for LWIR, whose emissive thermal patterns have limited photometric correspondence with visible features. We presume that collecting datasets with three or more synchronized spectral bands (e.g., RGB–NIR–SWIR or RGB–NIR–SWIR–LWIR) would enable direct inter-infrared alignment objectives and may improve the consistency of the shared representation, particularly for LWIR.

\paragraph{Sub-optimal Sampling and Modality Convergence.}
As detailed earlier, our training pipeline employs a uniform round-robin sampling strategy that allocates equal gradient updates to each modality. However, this assumes a uniform difficulty across alignment tasks. The equal allocation of training cycles likely results in "over-fitting" the reflectance-based bands while leaving the thermal modality under-optimized. An adaptive, loss-driven sampling strategy, where the model dynamically prioritizes modalities with higher alignment error, could potentially bridge this performance gap without increasing the overall computational budget. We will study this approach in our future work.

\bibliographystyle{unsrtnat}
\bibliography{references}  

\end{document}